\def\year{2020}\relax
\documentclass[letterpaper]{article} %DO NOT CHANGE THIS
\usepackage{aaai18}  %Required
\usepackage{times}  %Required
\usepackage{helvet}  %Required
\usepackage{courier}  %Required
\usepackage{url}  %Required
\usepackage{graphicx}  %Required
\usepackage{amssymb}
\usepackage{todonotes}
\usepackage{subcaption} % for using subfigure
\usepackage{rotating}
\usepackage{stackengine}
\usepackage{amsmath}
\usepackage{bbm}
\usepackage{hyperref}
\hypersetup{
  colorlinks,
  urlcolor=blue,
  citecolor=black,
  linkcolor=black}
\usepackage{listings}
\usepackage[linesnumbered,ruled]{algorithm2e}
\graphicspath{{figs/}}

\usepackage{subcaption}
\usepackage{bm}
\usepackage{amsmath}

\newcommand{\E}{\mathbb{E}}

\newcommand{\Sspace}{\mathcal{S}}
\newcommand{\Ospace}{\mathcal{O}}
\newcommand{\Aspace}{\mathcal{A}}

\DeclareFixedFont{\ttb}{T1}{txtt}{bx}{n}{8} % for bold
\DeclareFixedFont{\ttm}{T1}{txtt}{m}{n}{8}  % for normal

\usepackage{color}
\definecolor{deepblue}{rgb}{0,0,0.5}
\definecolor{deepred}{rgb}{0.6,0,0}
\definecolor{deepgreen}{rgb}{0,0.5,0}
\newcommand\pythonstyle{\lstset{
    language=Python,
    basicstyle=\ttm,
    otherkeywords={self},             % Add keywords here
    keywordstyle=\ttb\color{deepblue},
    emph={MyClass,__init__},          % Custom highlighting
    emphstyle=\ttb\color{deepred},    % Custom highlighting style
    stringstyle=\color{deepgreen},
    commentstyle=\ttm\color{olive},
    showstringspaces=false            % 
}}
\lstnewenvironment{python}[1][]
{
\pythonstyle
\lstset{#1}
}
{}
\newcommand\pythoninline[1]{{\pythonstyle\lstinline!#1!}}

\nocopyright  % Comment this out to show aaai copyright!

\begin{document}
% USE THIS FOR PUBLICATION
\title{\texttt{pomdp\_py}: A Framework to Build and Solve POMDP Problems}
\author{Kaiyu Zheng\thanks{corresponding author}\qquad Stefanie Tellex\\\text{Department of Computer Science,}\\\text{Brown University}\\\text{Providence, RI, USA}\\{\texttt{\{kzheng10, stefie10\}@cs.brown.edu}}
% <-this % stops a space
  % <-this % stops a space
}% <-this % stops a space
% \title{\pomdppy: A Framework to Build and Solve POMDP Problems}
% \author{Anonymous Authors% <-this % stops a space <-this % stops a space
% }% <-this % stops a space

\maketitle

% %%%%%%%%%%%%%%%%%%%%%%%%%%%%%%%%%%%%%%%%%%%%%%%%%%%%%%%%%%%%%%%%%%%%%%%%%%%%%%%%
%1st sentence should be a general statement of the problem's importance.  2nd sentence a one-sentence statement about the shortcomings of related work.  Next few sentences should describe our technical solution to the problem.  Last two or three sentences describe the evaluation.  Usually one sentence about our quantitative evaluation, and one sentence about our robot demonstration. 
\begin{abstract}
 In this paper, we present \texttt{pomdp\_py}, a general purpose Partially Observable Markov Decision Process (POMDP) library written in Python and Cython. Existing POMDP libraries often hinder accessibility and efficient prototyping due to the underlying programming language or interfaces, and require extra complexity in software toolchain to integrate with robotics systems. \texttt{pomdp\_py} features simple and comprehensive interfaces capable of describing large discrete or continuous (PO)MDP problems. Here, we summarize the design principles and describe in detail the programming model and interfaces in \texttt{pomdp\_py}. We also describe intuitive integration of this library with ROS (Robot Operating System), which enabled our torso-actuated robot to perform object search in 3D.  Finally, we note directions to improve and extend this library for POMDP planning and beyond.

\end{abstract}
\section{Introduction}

Partially Observable Markov Decision Processes (POMDP) are a sequential decision-making framework suitable to model many robotics problems, from localization and mapping \cite{ocana2005indoor} to human-robot interaction \cite{whitney2017reducing}. Early efforts in developing tools for POMDPs attempt to separate solvers from domain description by creating specialized file formats to specify POMDPs \cite{pomdpfileformat,pomdpxformat}, which are not designed for large and complex problems. Among libraries under active development, Approximate POMDP Planning Toolkit (APPL) \cite{somani2013despot} and AI-Toolkit \cite{aitoolbox} are implemented in C++ and contain numerous solvers. However, the learning curve for these libraries is steep as C++ is less accessible to current researchers in general compared to Python \cite{virtanen2020scipy}. POMDPs.jl \cite{egorov2017pomdps} is a POMDP library with a suite of solvers and domains, written in Julia. Though promising, Julia has yet to achieve a wide recognition and creates language barrier for many researchers. POMDPy \cite{emami2015pomdpy} is implemented purely in Python. Yet with an original focus on POMCP implementation, it assumes a blackbox world model in its POMDP interface, limiting its extensibility. Finally, a promising toolchain is to use Relational Dynamic Influence Diagram Language (RDDL) \cite{sanner2010relational} to describe factored POMDPs and solve them via ROSPlan \cite{cashmore2015rosplan}, recently demonstrated for object fetching \cite{canal2019probabilistic}. Nevertheless, using this set of tools adds overhead of using a classical fluent-based planning paradigm, which is not required to describe and solve POMDPs in general.

\begin{figure}[t]
    \centering
    \includegraphics[width=\linewidth]{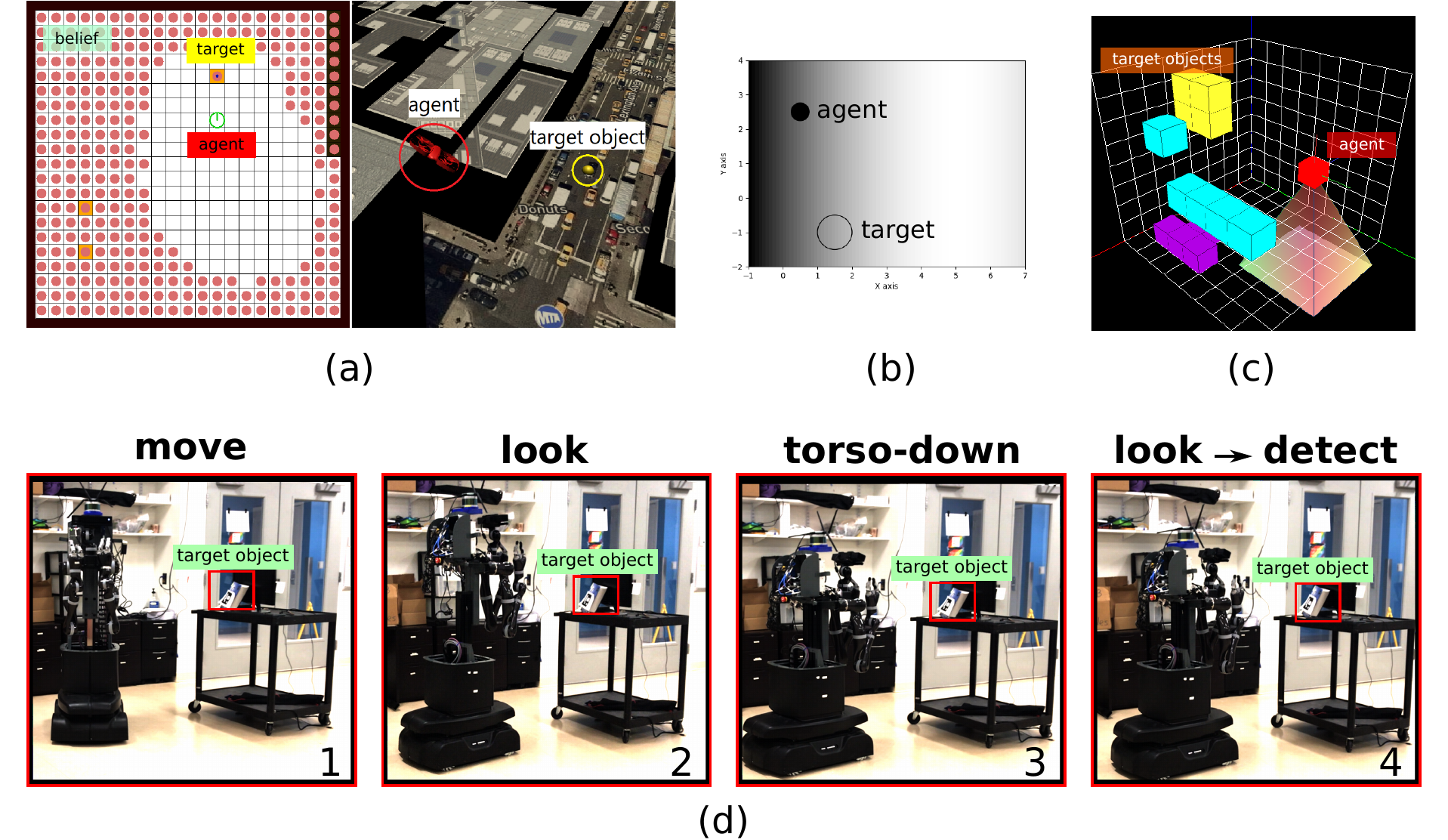}
    \caption{Example tasks implemented using \texttt{pomdp\_py}. (a) Object search in 2D with a simulated drone in Unity controlled using ROS{\protect\footnotemark}. (b) the light-dark domain with continuous spaces. (c) Object search in a 3D simulation with frustum-shaped field-of-view. (d) Object search in 3D implemented on a torso-actuated robot controlled using ROS; (d1-d4) shows a sequence of actions planned using an PO-UCT implemented in \texttt{pomdp\_py}, where the robot decides to lower its torso to search, and finds the object on the table. }
    \label{fig:firstfig}
\end{figure}
\footnotetext{We thank
Rebecca Mathew for kindly providing this figure.}

This leads to our belief that there lacks a POMDP library with simple interfaces that brings together both accessibility and performance. We address this demand by presenting \texttt{pomdp\_py}, a framework to build and solve POMDP problems written in Python and Cython \cite{behnel2011cython}. It features simple and comprehensive interfaces to describe POMDP or MDP problems, and can be integrated with ROS \cite{quigley2009ros} intuitively through \texttt{rospy}. In the rest of this paper, we first review POMDPs, then illustrate the design principles and key features of \texttt{pomdp\_py}, including integration with ROS. Finally, we note directions to improve and extend this library, in hope of cultivating an open-source community for POMDP-related research and development. The documentation of \texttt{pomdp\_py} is available at: \url{https://h2r.github.io/pomdp-py/html/}. Tutorials on example domains can be found in the documentation. Figure~\ref{fig:firstfig} shows several different domains that are implemented on top of the \texttt{pomdp\_py} framework. This library is currently actively developed as we continue our POMDP-related research.

\section{POMDPs}
\begin{figure}[t]
    \centering
    \includegraphics[width=\linewidth]{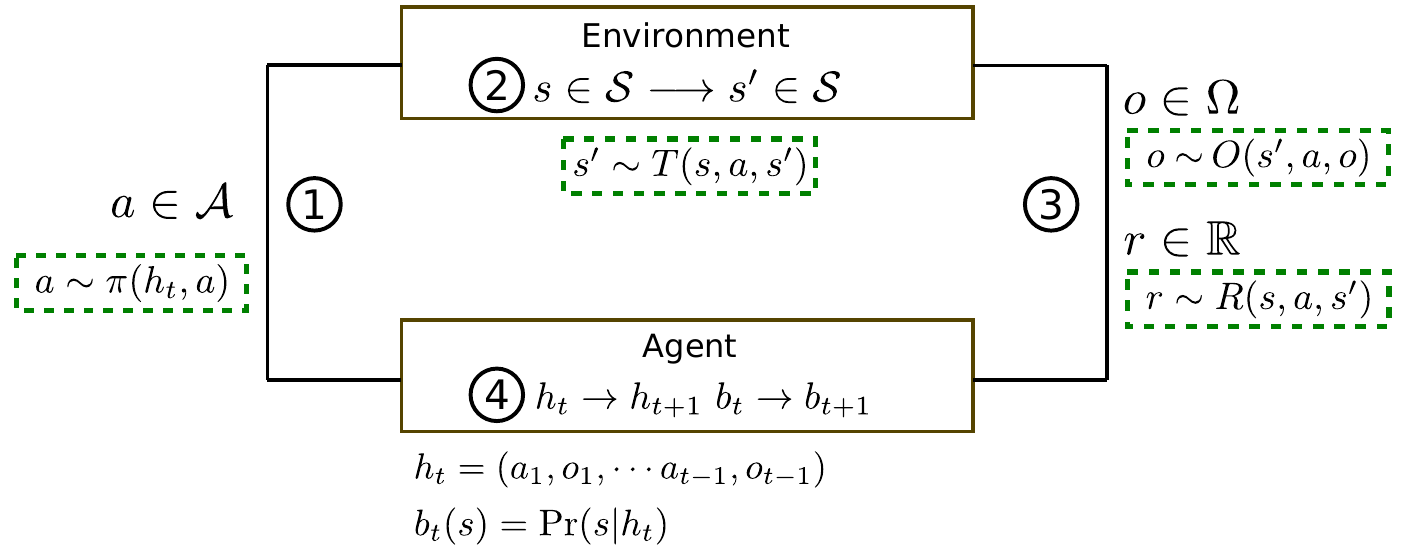}
    \caption{POMDP model of agent-environment interaction. (1)~Agent takes an action. (2)~Environment state transitions. (3)~Agent receives an observation and a reward signal. (4)~Agent updates history and belief.}
    \label{fig:pomdp}
\end{figure}

POMDPs \cite{kaelbling1998planning} model sequential decision making problems where the agent must act under partial observability of the environment state (Figure~\ref{fig:pomdp}). POMDPs consider both uncertainty in action effect (i.e. transitions) and observations, which are usually incomplete and noisy information related to the state. A POMDP is defined as a tuple $\langle\Sspace, \Aspace, \Ospace, T, O, R,\gamma\rangle$. The problem \emph{domain} is specified by $\Sspace$, $\Aspace$, $\Ospace$: the state, action, and observation spaces. At each time step, the agent decides to take an action $a\in\mathcal{A}$, which may be sampled from $a\sim\pi(h_t,\cdot)$ according to a \emph{policy model} $\pi(h_t,a)=\Pr(a|h_t)$. This leads to state change from $s$ to $s'\sim T(s,a,s')$ according to the \emph{transition model} $T$. Then, the agent receives an observation $o \sim O(s',a,o)$ according to the \emph{observation model} $O$, and reward $r\sim R(s,a,s'), r\in\mathbb{R}$ according to the \emph{reward model} $R$. Upon receiving $o$ and $r$, the agent updates its history $h_t$ and belief $b_t$ to $h_{t+1}$ and $b_{t+1}$. 
% In addition, the definition includes probabilistic \emph{models}: $T(s,a,s')=\Pr(s'|s,a)$ is the transition model that specifies the dynamics of the environment state; $O(s',a,o)=\Pr(o|s',a)$ is the observation model that specifies the uncertainty over an observation; $R(s,a,s')$ (also defined as $R(s,a)$) is the reward model that specifies reward feedback from the environment.
The goal of solving a POMDP is to find a policy $\pi(h_t,\cdot)$ which maximizes the expectation of future discounted rewards:
$
{V^{\pi}}(h_t)=\E\left[\sum_{k=t}^{\infty}\gamma^{k-t}R(s_k,a_k)\ \Big|\ a_k=\pi(h_k,\cdot)\right]
$, where $\gamma$ is the discount factor.
In \texttt{pomdp\_py}, a few key interfaces are defined to help organize the definition of POMDPs in a simple and consistent manner. \\

\noindent\textbf{Solvers.} Most recent POMDP solvers are \emph{anytime algorithms}~\cite{zilberstein1996using,ross2008online}, due to the intractable computation required to solve POMDPs exactly \cite{madani1999undecidability}. 
%This means that the solver can be stopped at any time and an approximate solution would be produced. The quality of the solution improves as the time budget increases.
There are currently two major camps of anytime solvers, point-based methods \cite{kurniawati2008sarsop,shani2013survey} which approximates the belief space by a set of reachable $\alpha$-vectors, and Monte-Carlo tree search-based methods \cite{silver2010monte,somani2013despot} that explores a subset of future action-observation sequences.

Currently, \texttt{pomdp\_py} contains an implementation of POMCP and PO-UCT \cite{silver2010monte}, as well as a naive exact value iteration algorithm without pruning \cite{kaelbling1998planning}. The interfaces of the library support implementation of other algorithms; We hope to cultivate a community to implement more solvers or create bridges between \texttt{pomdp\_py} and other libraries.\\

\noindent\textbf{Belief representation} The partial observability of environment state implies that the agent has to maintain a posterior distribution over possible states \cite{thrun2005probabilistic}. The agent should update this belief distribution through new actions and observations. The exact belief update is given by $b_{t+1}(s')=\eta\Pr(o|s',a)\sum_{s}\Pr(s'|s,a)b_t(s)$, where $\eta$ is the normalizing factor.
Hence, a naive tabular belief representation requires nested iterations over the state space to update the belief, which is computationally intractable in large domains. Particle belief representation is a simple and scalable belief representation which is updated through matching simulated and real observations exactly \cite{silver2010monte}. Different schemes of weighted particles have been proposed to handle large or continuous observation spaces where exact matching results in particle depletion \cite{sunberg2018online,garg2019despot}.

\texttt{pomdp\_py} does not commit to any specific belief representation. It provides implementations for basic belief representations and update algorithms, including tabular, particles, and multi-variate Gaussians, but more importantly  allows the user to create their own new or problem-specific representation, according to the interface of a generative probability distribution. 

\section{Design Philosphy}

\label{sec:design}
 Our goal is to design a framework that allows simple and intuitive ways of defining POMDPs at scale for both discrete and continuous domains, as well as solving them either through planning or through reinforcement learning. In addition, we implement this framework in Python and Cython to improve accessibility and prototyping efficiency without losing orders of magnitude in performance \cite{behnel2011cython,smith2015cython}. %Having a framework of such clarifies POMDP problems and enables modeling a wide range of robotics problems (and beyond), from localization to human-robot interaction.
% We then implement this framework in Python, a popular language behind numerous advances in artificial intelligence nowadays. We hope that this makes POMDP-related research or projects accessible to more people, and fosters a community where experiences are shared.
We summarize the design principles behind \texttt{pomdp\_py} below:
 \begin{itemize}
     \item Fundamentally, we view the POMDP scenario as the interaction between an \emph{agent} and the \emph{environment}, through a few important generative probability distributions ($\pi$, $T,O,R$ or blackbox model $G$). 
     \item The agent and the environment may carry different models to support learning, since for real-world problems especially in robotics, the agent generally does not know the true transition or reward models underlying the environment, and only acts based on a simplified or estimated model.
     \item The POMDP domain could be very large or continuous, thus explicit enumeration of elements in the spaces should be optional.
     \item The representation of belief distribution is decided by the user and can be customized, as long as it follows the interface of a generative distribution.
     \item Models can be reused across different POMDP problems. Extensions of the POMDP framework to, for example, decentralized POMDPs, should also be possible by building upon existing interfaces.
 \end{itemize}
 
 \section{Programming Model and Features}
\begin{figure}[t]
    \centering
    \includegraphics[width=0.96\linewidth]{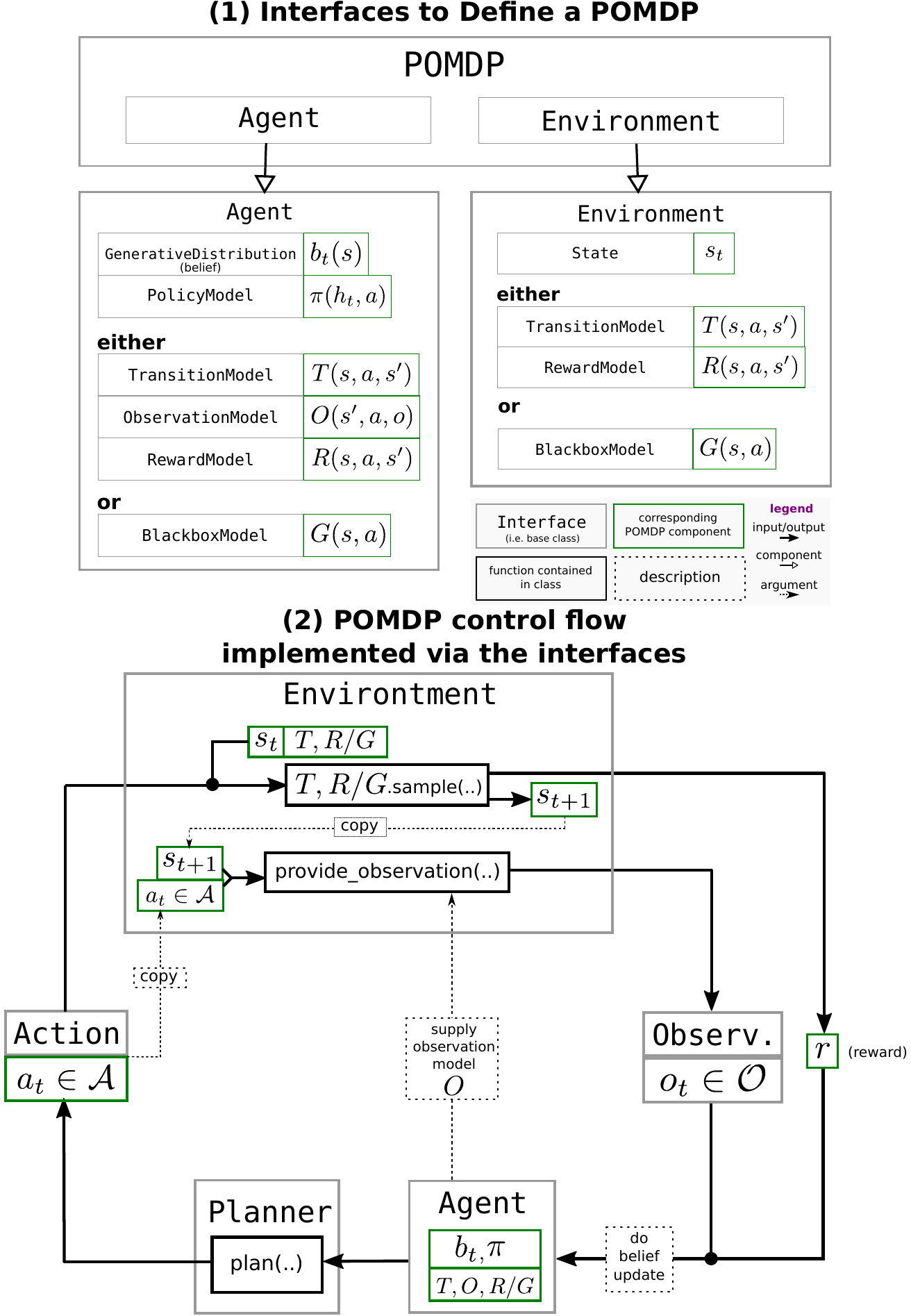}
    \caption{(1) Core Interfaces in the \texttt{pomdp\_py} framework; (2) POMDP control flow implemented through interaction between the core interfaces.}
    \label{fig:framework}
\end{figure}

 The basis of \texttt{pomdp\_py} is a set of simple interfaces that collectively form a framework for building and solving POMDPs. Figure~\ref{fig:framework} illustrates some of the key components of the framework and the control flow.

 When defining a POMDP, one first define the \emph{domain} by implementing the \texttt{State}, \texttt{Action}, \texttt{Observation} interfaces. The only required functions for each interface are \texttt{\_\_eq\_\_} and \texttt{\_\_hash\_\_}. For example, the interface for \texttt{State} is simply\footnote{Note that the code snippets here are modified or shortened slightly for display purposes. Please refer to the code on github: \url{https://github.com/h2r/pomdp-py/}}:
 \begin{python}
class State:
   def __eq__(self, other):
       raise NotImplementedError
   def __hash__(self):
       raise NotImplementedError      
 \end{python}
 Next, one define the \emph{models} by implementing the interfaces \texttt{TransitionModel}, \texttt{ObservationModel}, etc. (see Figure~\ref{fig:framework} for all). Note that one may define a different transition and reward model for the agent than the environment (e.g. for learning). One also defines a \texttt{PolicyModel} which (1) determines the action space at a given history or state, and (2) samples an action from this space according to some probability distribution.  Implementing any of them largely means to implement the \texttt{probability}, \texttt{sample} and \texttt{argmax} functions. For example, the interface for \texttt{ObservationModel}, modeling $O(s',a,o)=\Pr(o|s',a)$, is:
 %https://tex.stackexchange.com/questions/63729/how-to-use-mathematical-symbols-in-listing/63731
 \begin{python}[mathescape=true]
class ObservationModel:
   def probability(self, observation, next_state,
                   action, **kwargs):
       """Returns the probability Pr(o|s',a)."""
       raise NotImplementedError
   def sample(self, next_state, action, **kwargs):
       """Returns a sample o ~ Pr(o|s',a)."""
       raise NotImplementedError
   def argmax(self, next_state, action, **kwargs):
       """Returns o* = argmax_o Pr(o|s',a)."""
       raise NotImplementedError
   def get_all_observations(self, *args, **kwargs):
       """Returns a set of all possible
       observations, if feasible."""
       raise NotImplementedError        
 \end{python}
 It is up to the user to choose which subset of these functions to implement, depending on the domain. These interfaces aim to remind users the essence of models in POMDPs.

 To instantiate a POMDP, one provides parameter for the models, the initial state of the environment, and the initial belief of the agent. For the Tiger problem\footnote{\url{https://h2r.github.io/pomdp-py/html/examples.tiger.html}} \cite{kaelbling1998planning}, for example,
 \begin{python}
 s0 = random.choice(list(TigerProblem.STATES))
 b0 = pomdp_py.Histogram({State("tiger-left"): 0.5,       
         State("tiger-right"): 0.5})
 tiger_problem = TigerProblem(..., s0, b0)
 \end{python}
 Here, \texttt{TigerProblem} is a \texttt{POMDP} whose constructor takes care of initializing the \texttt{Agent} and \texttt{Environment} objects, and is instantiated by parameters (omitted), initial state and belief. Note that it is entirely optional to explicitly define a problem class such as \texttt{TigerProblem} in order to program the POMDP control flow, discussed below.
 
 To solve a POMDP with \texttt{pomdp\_py}, here is the control flow one should implement that contains the basic steps (see also Figure~\ref{fig:framework} for illustration):
\begin{enumerate}
    \item Create a planner (\texttt{Planner}), i.e. a POMDP solver.
    \item Agent plans an action $a\in\Aspace$ through the planner.
    \item Environment state transitions $s_t\rightarrow s_{t+1}$ according to its transition model.
    \item Agent receives an observation $o_t$ and reward $r_t$ from the environment.
    \item Agent updates history and belief. $h_t,b_t\rightarrow h_{t+1},b_{t+1}$, where $h_{t+1}=h_{t}(a_t,o_t)$.
    \item Unless \emph{termination condition} is true, repeat steps 2-5.
\end{enumerate}
The \texttt{Planner} interface is as follows. The planner may be updated given real action and real observation, which is necessary for MCTS-based solvers.
\begin{python}
class Planner:
    def plan(self, agent):    
        """The agent carries the information:
        Bt, ht, O,T,R/G, pi, needed for planning"""
        raise NotImplementedError
    def update(self, agent, action, observation):
        """Updates the planner based on real action
        and observation. Updates the agent belief 
        accordingly if necessary. """
        pass
\end{python}

\noindent\textbf{Code Organization.}  In a more complicated problem such as the Light-Dark domain \cite{platt2010belief} or Multi-Object Search with fan-shaped sensors \cite{wandzel2019multi}, it may be tricky to organize the code base and be consistent across different problems. Below we provide a recommendation of the package structure to use \texttt{pomdp\_py} to guide the development and facilitate code sharing:
\lstset{
  frame=tb,
  language=C,
  basicstyle={\ttm},
  keywordstyle={\ttb\color{deepblue}},
  commentstyle=\color{deepgreen}
}
\begin{lstlisting}
 - domain/
    - state.py        // State
    - action.py       // Action
    - observation.py  // Observation
    - ...             
 - models/
    - transition_model.py   // TransitionModel
    - observation_model.py  // ObservationModel
    - reward_model.py       // RewardModel
    - policy_model.py       // PolicyModel
    - ...
- agent/
    - agent.py  // Agent
    - ...
- env/
    - env.py  // Environment
    - ...
 - problem.py  // POMDP
\end{lstlisting}
The recommendation is to separate code for domain, models, agent and environment, and have simple generic filenames. As in the above tree, files such as \texttt{state.py} or \texttt{transition\_model.py} are self-evident in their role. The \texttt{problem.py} file is where the specific implementation of the \texttt{POMDP} class is defined, and where the logic of control flow is implemented. Refer to the Multi-Object Search example in the documentation for more detail\footnote{\url{https://h2r.github.io/pomdp-py/html/examples.mos.html}}.\\

\noindent\textbf{Object-Oriented POMDPs.}
OO-POMDP \cite{wandzel2019multi} is a particular kind of factored POMDP that factors the state and observation spaces into a set of $n$ objects. For instance, $\Pr(s'|s,a)=\prod_i\Pr(s'_i|s,a)$, $i\in\{1,\cdots,n\}$. The belief space is also factored, which allows the belief space to grow linearly instead of exponentially as the number of objects increases. Each object is of a certain class and has a set of attributes. The values of these attributes constitute the state of an object. In \texttt{pomdp\_py}, we provide interfaces to implement OO-POMDPs, which serves as an example of extending the basic POMDP framework to create another class of model. These interfaces include \texttt{OOState}, \texttt{OOBelief}, \texttt{OOTransitionModel}, etc.\\

\noindent\textbf{Integration with ROS.} 
ROS \cite{quigley2009ros} is an open-source system that builds a network connecting computing stations and robots, where \emph{nodes} interact with one another through publishing messages or making service requests. It is typical to separate nodes that manage resources and controls the robot from nodes that runs sophisticated algorithms. This is the case of \texttt{pomdp\_py} as well. The POMDP-related computations can be done on a node that implements the POMDP control flow (see the six steps above). Inside this node, when an action is selected by the \texttt{Planner} (step 1), the node can publish a message to the nodes for robot control so that the robot can execute the action (step 2). The environment state automatically updates in the real world as a result of that action (step 3), and the node receives the sensor measurements or other forms of observations through subscribed topics (step 4), and performs belief update (step 5). This process is repeated until termination condition is met (step 6). ROS provides a package \texttt{rospy} which eases the integration of the POMDP control flow with the robot system.\\

\noindent\textbf{3D Object Search with Torso-Actuated Robot.} We developed a novel approach to model and solve an OO-POMDP for the task of multi-object search in 3D. Using \texttt{pomdp\_py}, we implemented this approach in a simulated environment, and on Kinova MOVO, a torso-actuated mobile manpulator platform controlled with ROS. Figure~\ref{fig:firstfig} shows a sequence of actions that lead to object detection. In this demonstration, each planning step has a time budget of 3 seconds.

%\paragraph{Performance} % Do a test of number of simulations in the PO-UCT algorithm for Tiger and RockSample. Compare between Python, Python + Cython, Python 3.5, Python 3.7.
\section{Conclusions \& Future Work}
We present a POMDP library, named \texttt{pomdp\_py}, that brings together accessibility to programmers through Python as well as performance through Cython, with an intuitive design and straightforward integration with ROS. The programming model is designed to encourage organized development and code sharing within a community. We believe \texttt{pomdp\_py} has potential to facilitate research besides POMDP planning, including reinforcement learning, transfer learning, and multi-agent systems. For example, mulitple \texttt{Agent} objects could be instantiated, and different \texttt{RewardModel} classes can be created to represent different tasks. Finally, we call for support to create bridges between \texttt{pomdp\_py} and other libraries to make use of existing algorithm implementations.

% Different \texttt{RewardModel} can be created 
% This includes (PO)MDP for reinforcement learning where the agent learns a \texttt{PolicyModel} (model-free) or a \texttt{TransitionModel} and an \texttt{ObservationModel} (model-based). Multi-Agent POMDPs as well as task transfer learning are also natural extensions, where multiple \texttt{Agent} objects could be instantiated, and different \texttt{RewardModel} classes can be created to represent different tasks. We also call for support to create bridges between \texttt{pomdp\_py} and other libraries to make use of existing algorithm implementations.
\bibliographystyle{aaai}
\bibliography{references}
\end{document}